\documentclass[journal]{IEEEtran}

\usepackage[numbers]{natbib}
\usepackage{subcaption}
\usepackage[justification=centering]{caption}
\usepackage{float} 
\usepackage{graphicx}
\usepackage{mwe}

\usepackage[utf8]{inputenc} % allow utf-8 input
\usepackage[T1]{fontenc}    % use 8-bit T1 fonts
\usepackage{hyperref}       % hyperlinks
\usepackage{url}            % simple URL typesetting
\usepackage{booktabs}       % professional-quality tables
\usepackage{amsfonts}       % blackboard math symbols
\usepackage{nicefrac}       % compact symbols for 1/2, etc.
\usepackage{microtype}      % microtypography

\usepackage{makeidx}

\usepackage{multirow}
\usepackage{multicol}
\usepackage{amsmath}
\usepackage{cleveref}
\usepackage{amssymb}
\usepackage{csquotes}
\usepackage{color}
\usepackage{mathtools}
\usepackage[ruled,vlined]{algorithm2e}
\usepackage{algorithmic}
\usepackage[normalem]{ulem}
\usepackage{hyperref}

\newcommand{\bx}{{\bf x}}
\newcommand{\by}{{\bf y}}
\newcommand{\bw}{{\bf w}}

\newcommand{\bbf}{{\bf f}}
\newcommand{\bb}{{\bf b}}
\newcommand{\bu}{{\bf u}}
\newcommand{\bA}{{\bf A}}
\newcommand{\bB}{{\bf B}}

\newcommand{\bI}{{\bf I}}
\newcommand{\bK}{{\bf K}}
\newcommand{\bk}{{\bf k}}

\newcommand{\bZero}{{\bf 0}}

\newcommand{\bX}{{\bf X}}

\newcommand{\bU}{{\bf U}}

\newcommand{\bSigma}{{\boldsymbol \Sigma}}
\newcommand{\bmu}{{\boldsymbol \mu}}

\newcommand{\bLambda}{{\boldsymbol \Lambda}}

\begin{document}

\let\WriteBookmarks\relax
\def\floatpagepagefraction{1}
\def\textpagefraction{.001}

%\begin{frontmatter}

\title{Adaptive Sparse Gaussian Process}


\author{\IEEEauthorblockN{Vanessa G\'omez-Verdejo\IEEEauthorrefmark{1}, Emilio Parrado-Hern\'{a}ndez\IEEEauthorrefmark{1},
Manel Mart\'inez-Ram\'on\IEEEauthorrefmark{2}, \IEEEmembership{Senior Member,~IEEE}. \\
\IEEEauthorblockA{\IEEEauthorrefmark{1}Signal Processing and Communications Department, University Carlos III de Madrid, Leganés, 28911, Madrid, Spain.}\\
\IEEEauthorblockA{\IEEEauthorrefmark{2}Electrical and Computer Engineering Department, The University of New Mexico, Albuquerque, 87106, NM, USA.}}
\thanks{Corresponding author: Vanessa Gómez-Verdejo (email: vanessag@ing.uc3m.es). This work is partially supported by grants PID2020-115363RB-I00 funded by MCIN/AEI/10.13039/50110001103, Spain, TED2021-132366B-I00 funded by MCIN/AEI/10.13039/501100011033 and by the ``European Union NextGenerationEU/PRTR'', EPSCoR OIA-1757207, funded by NSF, USA, and the King Felipe VI Endowed Chair of the UNM.
}}

\maketitle

\begin{abstract}
Adaptive learning is necessary for non-stationary environments where the learning machine needs to forget past data distribution. Efficient algorithms require a  compact model update to not grow in computational burden with the incoming data and with the lowest possible computational cost for online parameter updating. Existing solutions only partially cover these needs. Here, we propose the first adaptive sparse Gaussian Process (GP) able to address all these issues. We first reformulate a variational sparse GP algorithm to make it adaptive through a forgetting factor. Next, to make the model inference as simple as possible, we propose updating a single inducing point of the sparse GP model together with the remaining model parameters every time a new sample arrives. As a result, the algorithm presents a fast convergence of the inference process, which allows an efficient model update (with a single inference iteration) even in highly non-stationary environments. Experimental results demonstrate the capabilities of the proposed algorithm and its good performance in modeling the predictive posterior in mean and confidence interval estimation compared to state-of-the-art approaches.
\end{abstract}


\section{Introduction}

Signal processing applications \cite{tan2018digital} usually require online learning methods, i.e., algorithms capable of self-updating as new data arrive in the system. Moreover, in applications such as spectral estimation, signal modeling, adaptive filtering, or array processing, the signals to be processed are nonstationary (their distributions change over time), so we need algorithms capable of adapting to the new data distributions and forgetting the past. This is known as adaptive learning \cite{haykin2002adaptive, ingle2005statisical}.

In this sense, Bayesian models \cite{murphy2012machine} and in particular Gaussian Processes (GP) \cite{williams2006gaussian} seem to be an ideal solution since each time a new sample arrives the predictive posterior can be updated using the previous posterior as new prior of the model and multiplying it by the likelihood of the new data. This idea can be further extended to models which include hidden latent variables and require learning algorithms such as Expectation Maximization, providing what is known in the literature as online Variational Bayes \cite{Ghahramani2000}. The main problem of these approaches relies on the fact that the predictive posterior parameters usually depend on all the training data and therefore, as new samples arrive, its complexity tends to grow with $\mathcal{O}(N^3)$, $N$ being the number of samples, which makes them intractable if not combined with pruning schemes \cite{Vaerenbergh2012}. 

A possible solution to this problem is to use low computational cost GP versions \cite{Liu2020} or, more specifically, to resort to  Sparse GP (SGP) \cite{quinonero2005unifying}. In these approaches, the model solution only depends on $M << N$ points in the observation space, called inducing points. This avoids the complexity of the model growing as each new sample arrives. However, this strategy complicates the  model inference since, in each iteration, the algorithm has to learn and update the kernel and noise parameters and, additionally, the positions of these inducing points and their variational distribution if we resort to variational versions of these models \cite{titsias2009variational}.

Despite the number of online applications, there are few proposals for online GP able to provide, on the one hand, a compact solution or, in other words, whose complexity does not grow with the incoming training data, while, on the other hand, keep a low computational cost and capable of efficiently updating the parameters of the model with every new data. For example,  \cite{csato2002sparse} proposed a compact Bayesian online algorithm based on an approximation to the real predictive posterior with a computational cost of $\mathcal{O}(NM^2)$. The drawback of this approach is that, despite sequentially updating the subset of $M$ relevant data, the remaining model hyperparameters, as the kernel parameters and noise variance, keep fixed. 

Other approaches directly propose to use SGP models and exploit the structure of the resulting kernel matrices. For example,  \cite{evans2018scalable} approximates the kernel matrix using $K$ eigenfunctions to obtain a fast computation of the likelihood derivatives. A more recent work \cite{stanton2021kernel} introduces a structured kernel interpolation approach of order $K$ to reduce the computational cost of updating the predictive distribution to $\mathcal{O}(MK^2)$. This model has the limitation of having to predefine (and keep fixed) the set of inducing points. Besides, it has the additional drawback of being based on SGPs, so the obtained posterior is an approximation to the exact one, and its performance tends to be worse than the batch approach.

To overcome this last limitation, it is preferable to rely on variational SGPs algorithms, where the fact of including a variational distribution over the inducing points avoids the error of approximating the posterior. In this line, we can find the works  \cite{hoang2015unifying} or \cite{cheng2016incremental} where the authors propose an incremental learning algorithm for variational  SGPs, although their hyperparameters are fixed during the training procedure.

Another interesting approach is the model presented in \cite{Streaming2017}, where the authors propose to update the variational bound with an online variational Bayesian scheme. This way the model is efficiently updated at each iteration with complexity similar to that of a standard SGP, but with the advantage of having a variational bound expression to be optimized with respect to all the model hyperparameters, including the inducing point locations.  

Within the models based on variational SGP formulations, we can find other methods \cite{hensman2013gaussian,Streaming2017} that use Variational Stochastic Inference (VSI) to obtain online versions, since the VSI strategy allows them to work with mini-batches of data, providing a straightforward scheme to add new data to the minibatch as they arrive. However, these approaches present two problems: first, the stochastic optimization assumes that the data subsampling process is uniformly random, an assumption that does not necessarily hold in online or non-stationary environments; second, this optimization usually requires that each mini-batch is processed in several iterations, which is often not compatible with the computational constraints of an online approach.

Finally, the major shortcoming that we have found in all the  above-mentioned algorithms for their use in real signal processing applications is that they are designed to work in an online environment, but not in nonstationary scenarios. To our knowledge, there are no adaptive versions of compact or SGP formulations.

So, to provide a solution to these needs, in this paper, we propose an adaptive GP that includes:
\begin{itemize}
\item Adaptive learning, or the ability to forget the information of the remote samples and, thus, the capability to self-adapting to non-stationary environments.

\item Efficiently  predictive distribution update with a cost of $\mathcal{O}(M^3)$ each time a new sample arrives.

\item If it is desired, at each iteration, it can update the set of inducing points and the rest of the hyperparameters of the model with a cost of $\mathcal{O}(NM^2)$. 

\item To update the inducing point set at each iteration, only a single inducing point is updated. This significantly reduces the number of parameters to be inferred, and a single inference iteration is usually sufficient to obtain an accurate value of the model parameters. 

\item Finally, unlike reference approaches, the proposed model can recover the solution of the  batch variational SGP formulation. That is, if at a given time instant we consider the same model parameters and eliminate the forgetting factor, its predictive distribution converges to that of the original model. As we will see in the experimental section, in practice, this translates into an improvement in the model performance.
\end{itemize}

The rest of the paper is structured as follows. Section 2 reviews GP models, paying special attention to SGPs and the variational version used as starting point of our proposal. Then, Section 3  introduces the proposed algorithm, starting by developing its adaptive formulation and then explaining how to do efficient online learning on this model. Section 4 analyzes the performance of the model showing over a load forecasting application the advantages of having a compact, efficient, and adaptive GP model able to track the signal changes over time. Finally, Section 5 presents the main conclusions of this work. 

The implementation of all proposed models is available at \url{   https://github.com/vgverdejo/AdaptiveSparseGP}.

\section{A review of Sparse Gaussian Processes}
\subsection{Gaussian Processes}
Assume a regression problem defined by a training dataset $\{\bx_n, y_n\}_{n=1}^N$, with $\bx_n \in \mathbb{R}^{D}$ and $y_n \in \mathbb{R}$, where each $y_n$ is generated by a noisy version of a latent function $f(\bx_n)=f_n$, i.e.
\begin{equation}\label{eq:Gaussian_model}
  y_{n}=f_n+\varepsilon_{n},~~\varepsilon_n\sim \mathcal{N}(0,\sigma^2).  
\end{equation}
Assume further that function $f_n$ can be expressed as 
\begin{equation}
    f_n=\langle \bw, \varphi(\bx_n) \rangle = \bw^\top \varphi(\bx_n),
\end{equation}
where  $\bw$ are the primal parameters, which are modeled as a multivariate Gaussian with zero mean and covariance matrix $\bSigma_p$. Function $\varphi(\cdot)$ is a nonlinear mapping of the input data $\bx_n$ into a Hilbert space $\mathcal{H}$ endowed with a dot product $\langle \varphi(\bx),\varphi(\bx')\rangle$, usually called a Mercer's kernel function, which, by virtue of the Mercer's Theorem \cite{Aizerman64}, is a dot product if and only if such function is positive semidefinite. From  expression \eqref{eq:Gaussian_model} the covariance between instances of the latent function can be computed as 
\begin{equation}
\begin{split}
    \mathbb{E}\left[f_n \cdot  f_m\right] = & \mathbb{E}\left[\varphi(\bx_n)^\top \bw \bw^\top \varphi(\bx_m)\right] \\ 
    = & \varphi(\bx_n)^\top \mathbb{E}\left[\bw \bw^\top\right] \varphi(\bx_m)\\
    = & \varphi(\bx_n)^\top \bSigma_p \varphi(\bx_m) \\
    = &  \langle\bSigma_p^{1/2}\varphi(\bx_n),\bSigma_p^{1/2}\varphi(\bx_m)\rangle =k(\bx_n, \bx_m).
    \end{split}
\end{equation}
Since $\bSigma_p$ is positive semidefinite, then function $k(\bx_n,\bx_m) $ is positive semidefinite and hence it is a valid kernel dot product.

The GP model for regression \cite{rasmussen2003gaussian} considers that the joint process of the latent functions $\bbf=[f(\bx_1),\ldots, f(\bx_N)]$ is drawn from a multivariate Gaussian prior $p(\bbf)$ with with zero mean and covariance $\bK_{xx}$, where this matrix contains all the kernel dot products $k(\bx_n, \bx_m)$ between input data $\bx_{n}$ and $\bx_{m}$ for $1\leq n,m\leq N$. From this model, we can obtain the predictive posterior for a test sample $\bx_*$ as
\begin{equation}
\begin{split}
        p(f_*|\by) & =\int p(f_*|\bbf)p(\bbf|\by)d\bbf = \\
        & = \mathcal{N}\left( f_*| \bk_{x*}^{\top}\left(\bK_{xx} + \sigma^2 \bI \right)^{-1} \by, \right.\\
        & ~~~~~~~~~\left. k_{**} - \bk_{x*}^{\top} \left(\bK_{xx} + \sigma^2 \bI \right)^{-1} \bk_{x*} \right),
\end{split}
\end{equation}
where $\bk_{x*}$ is a column vector with the kernel products $k(\bx_n, \bx_*)$, for $1\leq n\leq N$, $k_{**}=k(\bx^*,\bx^*)$ is the kernel product of $\bx_*$ with itself, and $\by= [y_1,\ldots, y_N] \in \mathbb{R}^{N}$ is the vector representation of all the training labels. The computation of this posterior requires the inversion of an $N\times N$ matrix, which takes a computational burden of $\mathcal{O}(N^3)$. 

\subsection{Sparse Gaussian Processes}

The  motivation of SGPs is to reduce this computational cost by including a set of $M$ inducing variables or inducing points $\bu_m$, $1\leq m\leq M$, that approximate the dual space to produce a compact GP model with computational burden $\mathcal{O}(NM^2)$. 

The first versions of the SGP can be unified in a common framework  \cite{quinonero2005unifying} considering that the covariance of the GP prior is modified by an approximate matrix. For example, in \cite{seeger2003fast} $\bK_{xx}$ is replaced by a Nystr\"on approximation $ \bK_{xu}\bK_{uu}^{-1}\bK_{ux}$, $\bK_{xu}$ and $\bK_{uu}$ being the kernel matrix of the inducing points with the training data and with themselves, respectively. An evolution of this method, introduced in \cite{snelson2005sparse}, corrects the approximated matrix with the term $\rm{diag}(\bK_{xx})- \rm{diag}(\bK_{xu}\bK_{uu}^{-1}\bK_{ux})$ so that the diagonal of $\bK_{xx}$ is exact.

The main drawback of these approximations is that they do not tend to the exact GP, since they start from an approximation to the real prior, so their posterior will be an approximation to the real one. Moreover, the inducing inputs constitute additional parameters to be inferred, increasing the overfitting risk. 

\subsection{Variational Sparse Gaussian Processes}
\label{sec:Standard_VSGP}

The Variational Sparse Gaussian Process (VSGP) introduced by Titsias in \cite{titsias2009variational} is intended to overcome the abovementioned limitations by minimizing the divergence between the exact GP posterior and a variational approximation  where the inducing points are modeled as variational parameters. This way, by minimizing the divergence with respect to these inducing points, we get the SGP to tend to the original one and, by introducing a variational prior over the inducing variables, we reduce the overfitting risk.  

To introduce this formulation, let us consider a set of inducing inputs $\bU=[\bu_1,\ldots, \bu_M]$, with their corresponding latent variables $\bbf_u=[f(\bu_1),\ldots, f(\bu_M)]^\top$. Then, the predictive posterior of the VSGP is given by
\begin{equation}
        p(f_*|\by) =\int \int p(f_*|\bbf_u,\bbf)p(\bbf|\bbf_u,\by)p(\bbf_u|\by)d\bbf d\bbf_u.
\end{equation}
Assuming that $\bbf_u$ is a sufficient statistic for $\bbf$, $ p(f_*|\bbf_u,\bbf)$ can be approximated by $p(f_*|\bbf_u)$. Moreover, approximating the posterior distribution $p(\bbf_u|\by)$ by a variational distribution $q(\bbf_u)$,  the following approximate posterior can be used 
\begin{equation}\label{eq:approximate_posterior}
\begin{split}
q(f_*) &=\int \int p(f_*|\bbf_u)p(\bbf|\bbf_u)q(\bbf_u)d\bbf d\bbf_u\\
&=\int p(f_*|\bbf_u)q(\bbf_u) d\bbf_u.
\end{split}
\end{equation}
Considering  that $q(\bbf_u)$ is a Gaussian distribution with mean $\bmu$ and covariance $\bA$, the following approximate predictive posterior is obtained
\begin{equation}\label{eq:q_f*}
\begin{split}
    q(f_*) & =  \mathcal{N}\left( f_*| m_* , v_* \right)\\
    m_*  &= \bk_{u*}^{\top} \bK_{uu}^{-1}\bmu\\
    v_* &= k_{**} -\bk_{u*}^{\top}  \bK^{-1}_{uu} \bk_{u*}+\bk_{u*}^{\top} \bK_{uu}^{-1}\bA\bK_{uu}^{-1}\bk_{u*},
\end{split} 
\end{equation}
where $\bk_{u*}$ is a column vector with the kernel products between $\bx_*$ and  $\bU$. The main advantage of this approach relies on the fact that once $\bmu$, $\bA$, and $\bU$ are obtained, the prediction of a new sample does not depend on training data, but only on these parameters. So the complexity of predicting a new sample is $\mathcal{O}(M^3)$ (it does not depend on $N$). 

To complete this model, one needs to find the variational parameters $\bmu$, $\bA$, and $\bU$. For this purpose,   \cite{titsias2009variational} introduces the following variational lower bound of the true log marginal likelihood 
\begin{equation}
   \log p(\by) \geq F_V(\bU, q(\bbf_u)) =
   \int q(\bbf_u) \log \frac{G(\bbf_u, \by)p(\bbf_u)}{q(\bbf_u)} d\bbf_u
   \label{eq:F_V_1}
   %= \int q(\bbf_u) \log G(\bbf_u, \by) + \log {\frac{p(\bbf_u)}{q(\bbf_u)}} d\bbf_u,
\end{equation}
where
\begin{equation}
\begin{split}
  \log G(\bbf_u, \by) & =  \log \mathcal{N} \left( \by | \bK_{xu} \bK_{uu}^{-1} \bbf_u, \sigma^2 \bI \right) \\
  &  - \frac{1}{2\sigma^2} \text{tr} \left\lbrace \bK_{xx} - \bK_{xu} \bK_{uu}^{-1} \bK_{ux}  \right\rbrace ,
\end{split}
\end{equation}
with $\text{tr}\left\lbrace \cdot \right\rbrace  $ being the trace operator. 
 To optimize \eqref{eq:F_V_1} w.r.t  the variational parameters, \cite{titsias2009variational} proposes to apply Jensen's inequality to move the logarithm out of the integral and, thus, to cancel distribution  $q(\bbf_u)$ to obtain a collapsed version of this bound, i.e.,
\begin{equation}\label{eq:F_V_2}
\begin{split}
\hspace{-0.2cm} F_V(\bU) &=  \log \mathcal{N} \left( \by | \bZero , \sigma^2 \bI + \bK_{xu} \bK_{uu}^{-1} \bK_{ux} \right) \\
  &  - \frac{1}{2\sigma^2} \text{tr} \left\lbrace \bK_{xx} -\bK_{xu} \bK_{uu}^{-1} \bK_{ux} \right\rbrace.
\end{split}
\end{equation}
and where $\bK_{xu} \bK_{uu}^{-1} \bK_{ux}$ is the covariance matrix of $p(\bbf_u)$.
This new bound can be maximized to obtain the optimal values of $\bU$, as well as other model hyperparameters, such as the kernel parameters and the noise variance. 

Finally, to obtain the optimal value of $q(\bbf_u)$, we can differentiate \eqref{eq:F_V_1} with respect to  $q(\bbf_u)$ and set it to zero. This leads to the fact that the optimal distribution is  proportional to $\mathcal{N} \left( \by | \bK_{xu} \bK_{uu}^{-1} \bbf_u, \sigma^2 \bI \right) p(\bbf_u)$ and identifying terms, we see that the optimal expression for $q(\bbf_u)$ is
\begin{equation}\label{eq:q_fu}
    \begin{split}
    q(\bbf_u) &= \mathcal{N}\left( \bbf_u|\bmu,\bA  \right)\\
        \bmu &= \sigma^{-2} \bK_{uu}\bB\bK_{ux}\by\\
        \bA &= \bK_{uu}\bB\bK_{uu},
    \end{split}
\end{equation}
where $\bB = \left( \bK_{uu} +\sigma^{-2} \bK_{ux} \bK_{xu}\right)^{-1}$. Eq. \eqref{eq:q_fu} allows to obtain $\bmu$ and $\bA$ to completely define the predictive posterior \eqref{eq:q_f*}. Note that both the optimization of \eqref{eq:F_V_2} and the computation of \eqref{eq:q_fu} have a computational burden  of $\mathcal{O}(NM^2)$. This provides fundamental advantages that allow novel and efficient adaptive GP variants, which will become apparent next.

\section{Adaptive Online VSGPs}\label{sec:adaptive_SGP}

In this section, we consider that we are working in a time-dependent framework, so in each time instant $t$ a new data pair $(\bx_t, y_t)$ arrives. To be able to efficiently deal with this new scenario, we next reformulate the previous VSGP model to make it able to:  (1) forget older samples (\emph{adaptive capabilities}); and, (2) be updated with the information of each new sample reusing the previous time instant model solution  (\emph{online learning}).

\subsection{Adaptive VSGP} \label{sect:adapt_SGP}
To endow the VSGP formulation with adaptive properties, we can modify the marginal likelihood bound \eqref{eq:F_V_1} by including a forgetting factor $\lambda$  (similar to that of the Recursive Least Squares filter \cite{haykin2002adaptive}) so that older data weigh less in the bound. Then, we can maximize this $\lambda$ dependent bound to obtain the model hyperparameters, as well as the new parameters of the $q(\bbf_u)$ distribution. 

For this purpose, as the only data-dependent term in $F_V(\bU, q(\bbf_u))$ is $\log G(\bbf_u, \by)$, we include $\lambda$ into this term and we start considering time instant $t$, so only data from $t'=1$ to $t$ are available, then,
\begin{equation}\label{eq:lambda_loglik}
\begin{split}
& \log G^{\lambda}(\bbf_u, \by) = \sum_{t'=1}^t  \lambda^{t-t'} \log G(\bbf_u, y_{t'})\\
& =  \sum_{t'=1}^t  \lambda^{t-t'}  \log \mathcal{N} \left( y_{t'} | \bk_{ut'}^{\top}  \bK_{uu}^{-1} \bbf_u, \sigma^2 \right) \\
&  - \frac{1 }{2\sigma^2}  \sum_{t'=1}^t  \lambda^{t-t'} \left( k_{t't'} -  \bk_{ut'}^{\top}  \bK_{uu}^{-1} \bk_{ut'} \right),  
\end{split}
\end{equation}
where $\by = [y_1,\ldots,y_t]^{\top}$, $\bk_{ut'}=\bk_{u\bx_{t'}}$ is a vector constructed with the kernel products between $\bU$ and sample $\bx_{t'}$, and $k_{t't'}$ is the kernel of $\bx_{t'}$ with itself. After some manipulations (see Appendix \ref{appendix:lambda_bound}), we can transform \eqref{eq:lambda_loglik} into
\begin{equation}
\begin{split}
  \log  G^{\lambda}(\bbf_u, \by) & \propto \log \mathcal{N} \left( \by | \bK_{xu} \bK_{uu}^{-1} \bbf_u, \sigma^2 {\bLambda}^{-1} \right) \\ 
      &- \frac{1}{2} \sum_{t'=1}^t  ( \lambda^{t-t'}-1) \log(2\pi\sigma^2) \\
      &  + \frac{1}{2 \sigma^2} \sum_{t'=1}^t  \lambda^{t-t'}\left( k_{t't'} -\bk_{ut'}^{\top} \bK_{uu}^{-1} \bk_{ut'} 
    \right),
\end{split}
\end{equation}
where $\bLambda $ is a $t\times t$ diagonal matrix with $\bLambda_{t't'} =  \lambda^{t-t'}$ and now $\bK_{xu}$ is constructed with the kernel products between samples $\bx_{1}, \ldots, \bx_{t}$ and the inducing points. The $\lambda$ dependent $F_V(\bU)$ bound is thus given by:
\begin{equation}\label{eq:lambda_bound}
\begin{split}
    F_V^{\lambda}(\bU) & = \log \mathcal{N} \left( \by | \bZero, \sigma^{2}\bLambda^{-1} + \bK_{xu} \bK_{uu}^{-1} \bK_{ux}  \right) \\ 
    & - \frac{1}{2}\sum_{t'=1}^t (\lambda^{t-t'}-1) \log(2\pi\sigma^2) \\
      &  + \frac{1}{2\sigma^2}\sum_{t'=1}^t \lambda^{t-t'}\left( k_{t't'} -\bk_{ut'}^{\top} \bK_{uu}^{-1} \bk_{ut'} 
    \right).
\end{split}
\end{equation}

The effect of the introduction of parameter $\lambda$ can be interpreted in Eq. \eqref{eq:lambda_bound} as an enhancement of the noise term that increases when $t'$ decreases. That is, an increasing uncertainty is attributed to the older training samples.  This is a parameter of the cost function, but not a parameter that can be included in the graphic model, and therefore, we do not optimize it maximizing the ELBO function with respect to $\lambda$, but it is a hyperparameter that must be selected by the user, as it is done in traditional adaptive algorithms. 

To complete this model, we obtain the adaptive version of the variational distribution of the inducing points, $q_{\lambda}(\bbf_u)$. For this purpose, its optimum value has to be proportional to $\mathcal{N} \left( \by | \bK_{xu} \bK_{uu}^{-1} \bbf_u, \sigma^2 \bLambda^{-1} \right) p(\bbf_u)$ and therefore its mean $\bmu_{\lambda}$ and covariance $\bA_{\lambda}$ are given by:
\begin{equation}\label{eq:q_tu_lamb}
    \begin{split}
        \bmu_{\lambda} &= \sigma^{-2} \bK_{uu}{\bB_{\lambda}} \bK_{ux} \bLambda \by\\
        \bA_{\lambda} &= \bK_{uu}{\bB_{\lambda}} \bK_{uu},
    \end{split}
\end{equation}
with $\bB_{\lambda} = \left( \bK_{uu} +\sigma^{-2} \bK_{ux} \bLambda \bK_{xu} \right)^{-1}$. Finally, by replacing the optimum values of $\bmu_{\lambda}$ and $\bA_{\lambda}$ into the predictive distribution of the VSGP (see \eqref{eq:q_f*}), we can obtain the mean and variance of adaptive predictive distribution as
\begin{equation} \label{eq:adap_pred_dist}
\begin{split}
   m_{\lambda,*} &= \sigma^{-2} \bk_{u*}^{\top} \bB_{\lambda}\bK_{ux} \bLambda \by \\
   v_{\lambda,*} &= k_{**} +\bk_{u*}^{\top}\left(\bB_{\lambda} - \bK^{-1}_{uu}\right)\bk_{u*}.
\end{split} 
\end{equation}

\subsection{Online update of the Adaptive VSGP}\label{subsec:online_update}

With the above model, we can train a GP for non-stationary environments so that the model will learn the distribution of the most recent samples. However, to make this model work efficiently in such scenarios, we need to be able to efficiently update its solution as new samples arrive. 

To carry out this online learning there are two possible working modes. Firstly, a fast implementation where noise and kernel parameters are considered fixed and, for every new data, we update the predictive distribution and, if needed, the inducing point set. Secondly, a more accurate solution where all model parameters are updated to efficiently track the data distribution changes.
Next, these approaches are explained in detail.

\subsubsection{Model update without inference over the model parameters}

If we consider that both the noise and kernel parameters are fixed, when a new training sample arrives, $(\bx_{t+1}, y_{t+1})$, we only have  to update the mean and variance of the adaptive predictive distribution (see Eq. \eqref{eq:adap_pred_dist}). Considering these values were $(m_{\lambda,*})_t$ and $(v_{\lambda,*})_t$, the new values at $t+1$ will be given by:
\begin{equation}\label{eq:update_pred_add}
\begin{split}
   (m_{\lambda,*})_{t+1} &= \sigma^{-2} \bk_{u*}^{\top}\left(\bB_{\lambda}\right)_{t+1} \left(\bK_{ux} \bLambda \by \right)_{t+1} \\
   (v_{\lambda,*})_{t+1} &= k_{**} + \bk_{u*}^{\top}\left(\left(\bB_{\lambda}\right)_{t+1} - \bK^{-1}_{uu}\right)\bk_{u*},
\end{split} 
\end{equation}
where
\begin{equation}\label{eq:data_add} 
\begin{split}
   \left(\bB_{\lambda}\right)_{t+1}  &= \left( \bK_{uu} +\sigma^{-2}\left( \lambda \left(\bK_{ux} \bLambda \bK_{xu}\right)_{t} +  \bk_{u,t+1}
   \bk_{u,t+1}^{\top} \right)\right)^{-1}\\ 
   \left(\bK_{ux} \bLambda \by \right)_{t+1} &=  \lambda \left(\bK_{ux} \bLambda \by \right)_{t} + \bk_{u,t+1} y_{t+1},
\end{split} 
\end{equation}
where $\bk_{u,t+1}$ is a column vector constructed with the kernel dot products between $\bU$ and sample $\bx_{t+1}$. 

Note that as we are working with an adaptive algorithm, the model can forget the past without explicitly removing the old data. Besides, the data-dependent terms, $\left(\bK_{ux} \bLambda \by \right)_{t}$ and $\left(\bK_{ux} \bLambda \bK_{xu}\right)_{t}$ do not increase its size when new data are added, so the predictive distribution complexity does not increase with each new data. 
However, if we want to limit the memory needed to store past samples (necessary, for example, to recompute $\bK_{ux}$ when the inducing points are updated), it is advisable to remove old data and keep a window of length $T$. This can be easily done during the addition of the data using these update rules, which leads to an expression alternative to those in Eq. \eqref{eq:data_add}:
\begin{equation}\label{eq:data_remove} 
\begin{split}
   \left(\bB_{\lambda}\right)_{t+1}  &= \left( \bK_{uu} +\sigma^{-2}\left( \lambda \left(\bK_{ux} \bLambda \bK_{xu}\right)_{t} + \bk_{u,t+1}
   \bk_{u,t+1}^{\top} \right) \right. \\
   & - \left.\lambda^{T} \bk_{u,t-T}
   \bk_{u,t-T}^{\top} \right)^{-1}  \\
   \left(\bK_{ux} \bLambda \by \right)_{t+1} &= \lambda \left(\bK_{ux} \bLambda \by \right)_{t} + \bk_{u,t+1} y_{t+1}  - \lambda \bk_{u,t-T} y_{t-T}.
\end{split} 
\end{equation}
If the values of $\left(\bK_{ux} \bLambda \bK_{xu}\right)_{t} $ from the previous iteration are saved, the complexity of these operations is $O(M^3)$ and $O(M)$, respectively.

In non-stationary scenarios, it is expected that as new data arrive, their distribution changes and we also need to update the positions of the inducing points to be able to represent the support of the new distribution. To avoid applying a computationally expensive inference process to update the position of all inducing points, when the inducing set is not representative enough, we propose to directly add the new data as a new inducing point. Besides, to limit the length of this set, we can remove the less representative inducing points.

To analyze the representativeness of the inducing set, we can use as a criterion the regularization term of the adaptive collapsed marginal likelihood \eqref{eq:lambda_bound}, i.e.,
\begin{equation} \label{eq:R_inducing_tot}
R_{\rm tot} = \sum_{t'=1}^t  \lambda^{t-t'}\left( k_{t't'} -   
\bk_{u,t'}^{\top} \bK_{uu}^{-1} \bk_{ut'} \right). 
\end{equation}
This quantity measures the $\lambda$ weighed error generated by predicting the training latent values from the inducing points. In fact, the non-adaptive version of this criterion has been already used by \cite{smola2000sparse, lawrence2002fast} for the selection of inducing points from the training data. So, we can analyze this quantity in each iteration, and in the case is larger than a given threshold ($R_{th,tot}$), we can decide to include the observation $\bx_{t+1}$ as a new inducing point. Besides, according to this criterion, we can consider that the relevance of each inducing point, $R_m$ for $m=1, \ldots, M$, is given by
\begin{equation} \label{eq:R_inducing}
R_m = \sum_{t'=1}^t  \lambda^{t-t'}  k_{mm}^{-1} k_{mt'}^2, 
\end{equation}
such that $R_{\rm tot} =\sum_{t'=1}^t  \lambda^{t-t'} k_{t't'}  - \sum_{m=1}^M R_m $. So, we can now set a relevance threshold ($R_{th}$) over $R_m $ to detect the useless inducing points and remove them. 

Finally, to complete this algorithm, we  need to update the predictive distribution again when the inducing point set is modified. For this purpose, given that there are $k$ elements in the inducing point set, we add a new inducing point $\bu_{k+1}$ to this set, so the new predictive mean and variance are:
\begin{equation}\label{eq:update_pred_inducing} 
\begin{split}
   (m_{\lambda,*})_{k+1} =& \sigma^{-2} \left( 
   \bk_{u,*} \right)_{k+1}^{\top} \left(\bB_{\lambda}\right)_{k+1} \left(\bK_{ux}\right)_{k+1} \bLambda \by  \\
   (v_{\lambda,*})_{k+1} =& k_{**} + \left( 
   \bk_{u,*} \right)_{k+1}^{\top}\cdot\\&\cdot \left(\left(\bB_{\lambda}\right)_{k+1} -  \left(\bK^{-1}_{uu}\right)_{k+1}\right) \left( \bk_{u*}\right)_{k+1}.
\end{split} 
\end{equation}

Updating $\left( \bk_{u,*}\right)_{k+1}$ and $\left(\bK_{ux}\right)_{k+1}$ is straightforward since it only implies adding a new element to this vector and a row  to the  matrix. The update of $\left(\bK^{-1}_{uu}\right)_{k+1}$ can be efficiently computed from  $\left(\bK^{-1}_{uu}\right)_{k}$ using the properties of the block matrix inversion \cite{meyer2000matrix}, if we consider  that  $\left(\bK_{uu}\right)_{k}$  is an update of $\left(\bK_{uu}\right)_{k+1}$  with a new row and column.
A similar procedure can be applied to $\left(\bB_{\lambda}\right)_{k+1}$ taking into account that
\begin{equation}\label{eq:update_B_inducing1}
\begin{split}
\left(\bB_{\lambda}\right)_{k+1} & = \left( \bK_{uu} +\sigma^{-2} \bK_{ux} \bLambda \bK_{xu} \right)^{-1} \\
& = 
\left( \begin{array}{ll} 
\left(\bB_{\lambda}^{-1} \right)_{k} & \left(\bb_{\lambda}\right)_{k+1} \\
\left(\bb_{\lambda}^{\top}\right)_{k+1} & \left(b_{\lambda}\right)_{k+1}
\end{array} \right)^{-1},
\end{split} 
\end{equation}
where
\begin{equation}\label{eq:update_B_inducing2}
\begin{split}
   \left(\bb_{\lambda}\right)_{k+1}  &= \bk_{u,k+1} + \sigma^{-2} \left(\bK_{ux}\right)_{k}  \bLambda \bk_{x,k+1}  \\
   \left(b_{\lambda}\right)_{k+1} & = k_{k+1,k+1} + \sigma^{-2}  \bk_{x,k+1}^{\top}   \bLambda \bk_{x,k+1}.
\end{split} 
\end{equation}
This way, the addition of a new inducing point has a computational cost of $\mathcal{O}(M^2)$. 

Note that for the computation of $\left(\bK_{ux}\right)_{k+1}$, we need to store matrix $\bX$ with all previous training data, although due to the $\lambda$ factor, only the most recent data influence over the solution; so, as we have argued before, we can keep a window of $T$ data to reduce the computational requirements without any performance degradation. 

When we need to remove an irrelevant inducing point for being useless or to keep the inducing point set with $M$ elements, we will have to incur a higher computational burden. This is because the inducing point to be removed occupies an arbitrary position in the matrix; therefore, a rank 1 update is not possible and we have to compute the inversions of  $\left(\bB_{\lambda}\right)_{k}$ and  $\left(\bK_{uu}\right)_{k}^{-1}$ from scratch with a computational cost of $\mathcal{O}(M^3)$ in the worst case that $k=M$. 

A summary of this fast model update, called fast-AGP, is included in Algorithm \ref{alg:AOVSGP_noInference}, where the model at the previous instant or a model initialized with some data history is considered as input.

\begin{algorithm}[t]
\footnotesize
\SetAlgoLined
\caption{Fast Adaptive VSGP (fast-AGP)}
\label{alg:AOVSGP_noInference}

$\rm{(fast-AGP)}_{t+1}$ = Update\_fast-AGP $\left( \rm{(fast-AGP)}_t,  \right.$ \\
~~~~~~~~~~~~~~~~~~~~~~~~~~~~~~~~~~~~~~~~$\left. (\bx_{t+1}, y_{t+1}), T, R_{th}, R_{th,tot} \right)$ \\

{\bf Inputs}: Previous model (fast-AGP)$_t$, new data sample $\bx_{t+1}$, $y_{t+1}$, length of data window (optional) $T$, and relevance thresholds to update the inducing point set  ($R_{th}$ and $R_{th,tot}$).

\tcp{Update predictive distribution with new data}
Compute new $(m_{\lambda,*})_{t+1}$ and $(v_{\lambda,*})_{t+1}$ with Eq.  \eqref{eq:update_pred_add} and \eqref{eq:data_add}. 

If we keep a data window of length $T$, use Eq. \eqref{eq:update_pred_add} and \eqref{eq:data_remove}.

\tcp{Add a new inducing point if it is needed}
If $R_{\rm tot} > R_{th,tot}$ (see Eq. \eqref{eq:R_inducing_tot}), set $\bu_{k+1} = \bx_{t+1}$ and update the predictive distribution with Eq. \eqref{eq:update_pred_inducing} -\eqref{eq:update_B_inducing2}.

\tcp{Remove useless inducing points}

For $m = 1, \ldots, k+1$, compute \eqref{eq:R_inducing} and remove inducing points with $R_m<R_{th}$ or those with the lowest $R_m$ so that the set length is no larger than $M$.

Then, if needed, update $\left(\bB_{\lambda}\right)$ and $\bK_{uu}^{-1}$and, later, $(m_{\lambda,*})_{k+1}$ and $(v_{\lambda,*})_{k+1}$.

\tcp{Return model fast-AGP model at $t+1$ }
Return $\rm{(fast-AGP)}_{t+1}$
\end{algorithm}

\subsubsection{Model update with inference over the model parameters} 
We can improve the performance of the previous algorithm by updating all the model parameters (kernel parameters, noise variance, and the position of the new inducing point) in each iteration.

To optimize these parameters, we need to maximize Eq. \eqref{eq:lambda_bound}, which leads to a computational burden of $\mathcal{O}(TM^2)$, considering that we limit the training data to a window of length $T$. In this case, since we will have to recalculate the kernel matrices every time we update the kernel parameters, we cannot save computation in the updates with rank one updates, so the predictive distribution will have to be calculated from scratch with a cost of $\mathcal{O}(TM^2)$. A summary of this approach is included in Algorithm \ref{alg:AOVSGP_Inference}.

\begin{algorithm}[t]
\footnotesize
\SetAlgoLined
\caption{Adaptive VSGP (AGP)}
\label{alg:AOVSGP_Inference}

$\rm{(AGP)}_{t+1}$ = Update\_AGP $ \left(\rm{(AGP)}_{t}, (\bx_{t+1}, y_{t+1}), T, R_{th}\right)$
\\

{\bf Inputs}: previous model (AGP)$_t$, new data sample $\bx_{t+1}$, $y_{t+1}$, length of data window (optional) $T$ and relevance threshold for inducing points  ($R_{th}$).

\tcp{Update the data window}
Include $(\bx_{t+1}, y_{t+1})$ to the model data set and remove $(\bx_{t-T}, y_{t-T})$.

\tcp{Remove useless inducing points and limit set size to $M-1$}
For $m = 1, \ldots, k+1$, compute \eqref{eq:R_inducing} and remove inducing points with $R_m<R_{th}$ or those with the lowest $R_m$ so that the set length is no larger than $M-1$.

\tcp{Add a new inducing point}
Set $\bu_{k+1} = \bx_{t+1}$ 

\tcp{Update model parameters}
Compute new values of $(\sigma^2)_{t+1}$, $(\bu_{k+1})_{t+1}$ and kernel parameters  $(\theta)_{t+1}$ by maximizing Eq. \eqref{eq:lambda_bound}.

\tcp{Update predictive distribution with new parameters}
Compute the predictive distribution at $t+1$ with \eqref{eq:adap_pred_dist}.

\tcp{Return model AGP model at $t+1$ }
Return $\rm{(AGP)}_{t+1}$ 
\end{algorithm}

Note that despite the need to compute the predictive function from scratch, the computational burden of this approach is significantly lower than one of the updates of a standard VSGP starting from the parameters optimized in the last time instant. This is because the number of parameters to update is significantly lower (we only update the position of one inducing point, not the whole set), so the convergence is likely to be achieved in a lower number of inference iterations; even, as we will show in the experimental section, a single iteration per time instant. Moreover, this procedure is subject to further simplifications. For example, the full parameter update may not be needed every time and, if we need to add a new inducing point, its point position can be set to that of a new data point, i.e., $\bu_{k+1} = \bx_{t+1}$; this way, no inference is needed in these iterations and the computational cost of the parameter update is lowered further.

\section{Experiments}\label{sec:experiments}

In this section, we evaluate the performance of the proposed adaptive models against several reference methods. For this purpose, we first consider a synthetic unidimensional problem where we can easily control the non-stationary environment to evaluate the advantages and drawbacks of the different methods. Second, we test these methods on a load forecasting application.

\subsection{Experimental setup}

To analyze the performance of the proposed model, we have considered the two introduced versions: 
\begin{itemize}
\item Adaptive VSGP without inference (fast-AGP) (see Algorithm \ref{alg:AOVSGP_noInference}), where only the set of inducing points is updated to represent the new data space, but the kernel parameters and the model noise are considered constant. Besides, the inducing point set is updated without any inference, since new data are included as new inducing points, and the useless ones are efficiently removed.
\item Adaptive VSGP with inference (AGP) (see Algorithm \ref{alg:AOVSGP_Inference}), where all model parameters are updated using a single inference iteration every time a new sample arrives.
\end{itemize}

In both cases, for automatic update of the set of inducing points, $R_{th,tot}$ at time instant $t$ is set as $\frac{1}{T}\sum_{t'=t-T}^t  \lambda^{t-t'} k_{t't'} $ and $R_{th} = r_{th} \max_{m}(R_m)$  where $r_{th}$ is a value which has been arbitrarily set to $10^{-4}$ in all the experiments below.

To analyze the performance of the proposed methods, we use the following reference approaches:
\begin{itemize}
    \item An adaptive VSGP where we use VSI (Variational Stochastic Inference) (AGP-VSI) to update the model parameters in each iteration (see Appendix \ref{sec:VSI_VSGP} for details of this formulation). For this model, we also consider a window of length $T$ so that only these $T$ samples are used for each data chunk. To make this algorithm converge, at each time instant, we need  $50$ iterations.
    
    \item Windowed VSGP (w-VSGP): This is the standard variational SGP of \cite{titsias2009variational} using the implementation of \cite{bingham2019pyro}, where we use a sliding window of length $T$ to simulate the adaptive scenario. As in the previous model, for each new data, we need to use $50$ inference iterations to adjust the model parameters.
    
    \item Online SGP (OSGP): this is the online approach proposed in \cite{Streaming2017}. In this case, instead of using the original implementation of the authors\footnote{https://github.com/thangbui/streaming\_sparse\_gp}, for comparison purposes, we have preferred to use its PyTorch implementation\footnote{https://github.com/wjmaddox/online\_gp/blob/main/online\_gp/models/streaming\_sgpr.py}. Here, we have used the learning rate parameters recommended by the authors ($lr=0.1$ for the noise and kernel parameters and $lr=0.2$ for the inducing points) with a Cosine Annealing optimizer. Here, as for the proposed methods, the online inference of the model uses a single iteration.
    
    \item The Kernel Interpolation for Scalable Online GP (WISKI) approach \cite{stanton2021kernel} using the authors implementation\footnote{Fhttps://github.com/wjmaddox/online\_gp}. For this model, since the inducing points are not updated by the model, we have had to predefine a grid over the input space to set them. Besides, as the definition of this grid raises computational problems in high-dimensional spaces when the input dimension is larger than two, we have projected the data into a two-dimensional space with a one-layer neural network (as the authors suggest); the weights of this net are learned with the remaining parameters of the model.

    The state of the art in online GP includes other methods such as those based on local approximations (see e.g. \cite{nguyen2008local,lederer2020real}) and sparse methodologies whose strategy is based on an online reduction of the kernel matrix size through similarity criteria (prominently, the work in \cite{koppel2021consistent}). The first methodology is not sparse and its implementation to our applications is not direct, since its computational complexity increases indefinitely with time and, therefore, its application to our experimental setup will involve a pruning procedure to bound such complexity. This is not proposed in the original paper. The second methodology is sparse, but the algorithms do not involve the optimization of a set of inducing points to be statistically significant for the corresponding posterior distribution. Therefore, we do not include these approaches in our comparisons.
    
\end{itemize}

For all the models and experiments, we used a square exponential kernel with its width and variance as parameters. To initialize the different models, we considered a window of length $T$ and we used the first $T$ samples to train a first model (in batch) during $200$ iterations. For the inference of all these models (either during their initialization or, if used, during online learning), except for OSGP which used the optimization proposed by the authors,  we applied an Adam algorithm with a learning rate of $0.05$.
For adaptive algorithms, the value of the forgetting factor ($\lambda$) was set as a function of the window length so that $ \lambda^{T} = 0.1 $. 

Experiments were run on a MacBook M1  with 32GB RAM.

\subsection{Toy example}

\begin{figure*}
    \centering
    \includegraphics[width=0.8\textwidth]{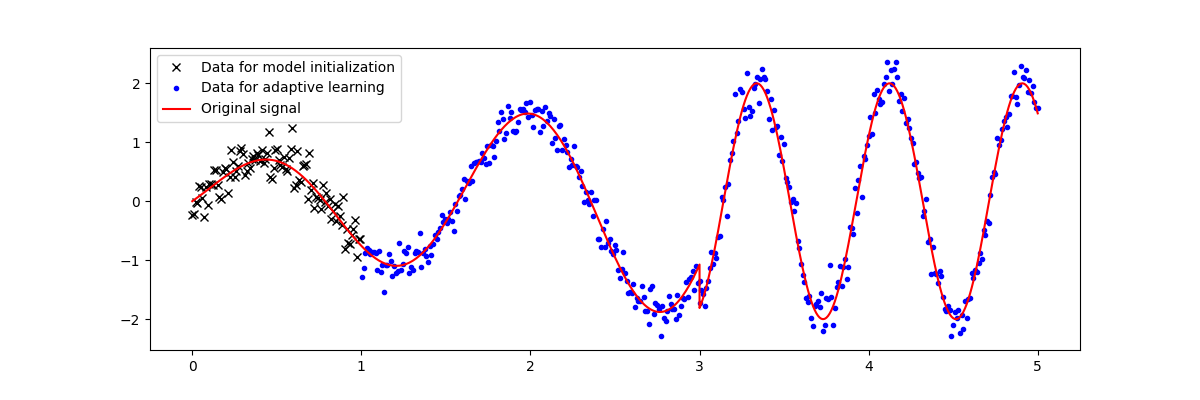}
    \caption{Example of the synthetic problem. Black points were used for the model initialization and blue samples for online learning.}
    \label{fig:toy_data}
\end{figure*}

As a first adaptive scenario, we generated a synthetic but challenging data set. In particular, we  generated a sinusoidal signal:
$$ y(t) = A_t {\rm sin}(2  \pi f t) + \epsilon_t$$
with $500$ samples uniformly distributed in the interval from $0$ to $5$. The first $300$ samples (in the interval $[0,3]$) use  $f = 4/2\pi$ with an increasing amplitude, $A_t$, from $0.5$ to $2$. The remaining $200$ samples consider $A_t = 2$ and a frequency of $8/2\pi$. In both cases, these sinusoidal signals were contaminated with a Gaussian noise of zero mean and standard deviation of $0.2$. In Figure \ref{fig:toy_data} we include an example of this data set.

For all models under study, we used a data window of $T=100$ samples, so we have trained a first model with first $100$, and we used $M=10$ inducing points; for WISKI model as we have to predefine an inducing points grid we set $40$ inducing points in the interval from $0$ to $5$. Finally, for the adaptive models, $\lambda$ was set to $\lambda = 0.97724$, so $\lambda^{100}\approx 0.1$. 

\begin{figure*}
    \centering
    \includegraphics[width=\textwidth]{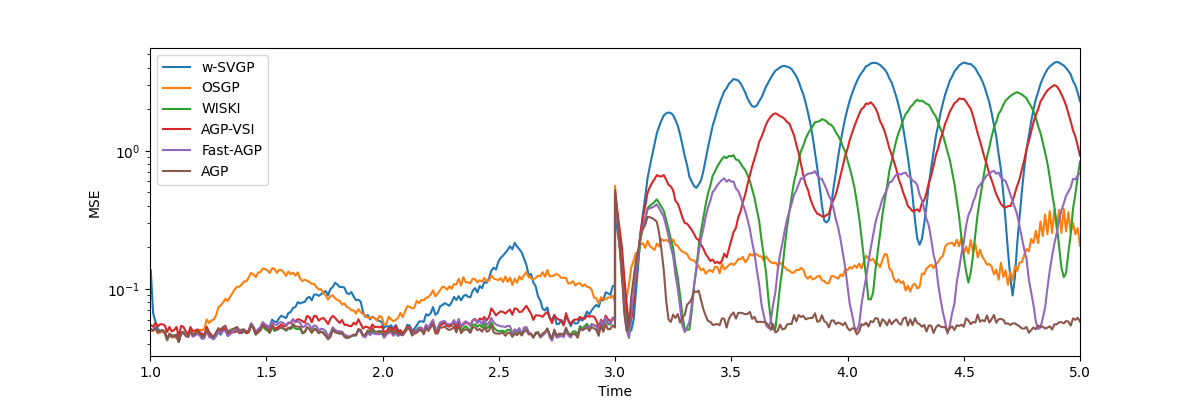} \\
    \caption{Evolution of the MSE during the online learning in the synthetic dataset.}
    \label{fig:MSE_toyproblem}
\end{figure*}

\begin{figure*}
    \centering
    \includegraphics[width=\textwidth, trim={2cm 1.5cm 2cm 0},clip ]{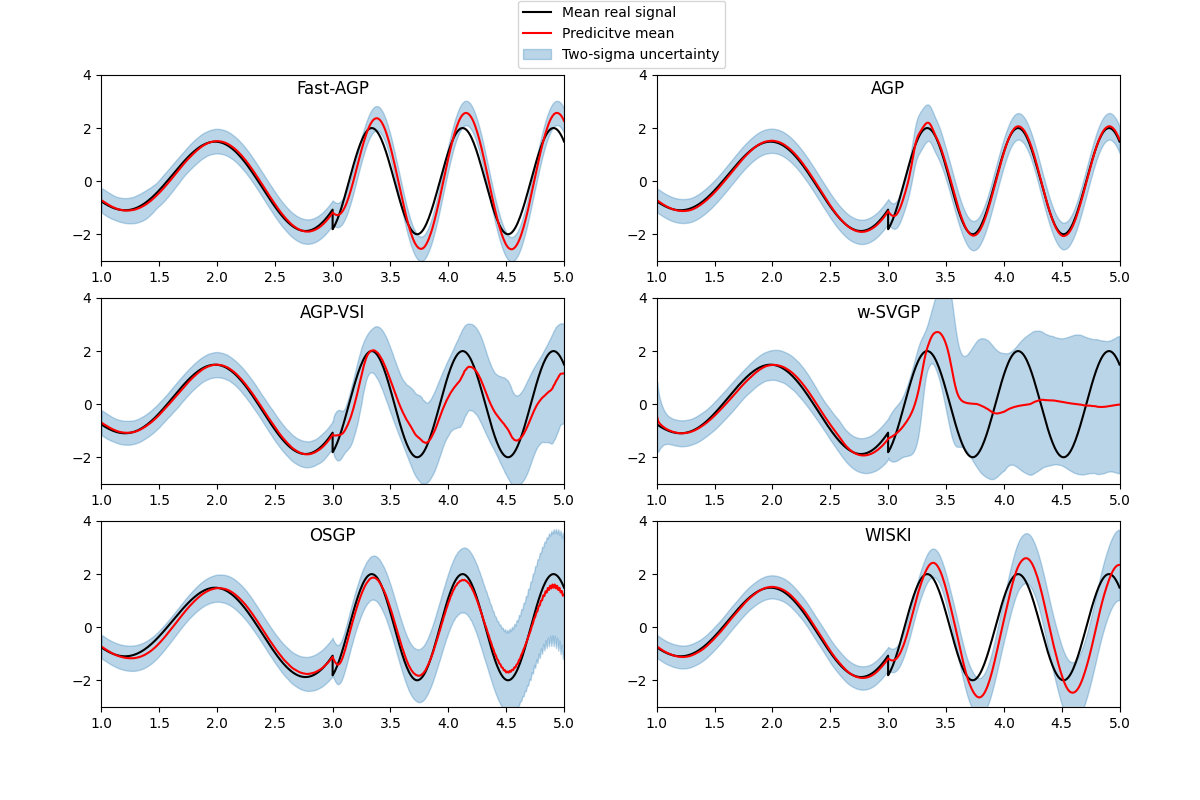} 
    \caption{Predictive distribution for the different methods under study along the online learning in the synthetic dataset.}
    \label{fig:Pred_dist_toyproblem}
\end{figure*}

To analyze the performance of the different methods, we  computed the Mean Square Error (MSE) and the percentage  of samples whose prediction error was less than $\pm 2 \left(v_{\lambda,*}  + \sigma^2\right)$ (see Eqs. \eqref{eq:update_pred_add} and \eqref{eq:update_pred_inducing}), known as $95\%$ confidence interval (95\% CI) when the model at the time instant $t$  (i.e, trained with samples from $t-T+1$ to $t$)  predicts the output at $t+1$. Besides, to obtain representative results, we  generated $1000$ different runs of the dataset and averaged the results for all the runs. In Table \ref{tab:results_toyproblem} we include these results for all the methods including their Training Times (TT). Besides, in Figures \ref{fig:MSE_toyproblem} and \ref{fig:Pred_dist_toyproblem} we include, respectively, the evolution of the MSE and the predictive distribution during the online learning of the algorithms.

\begin{table*}[t]
    \centering
    \begin{tabular}{l||c|c|c|c|c|c|}
&  w-SVGP & OSGP & WISKI  & AGP-VSI &  fast-AGP & AGP\\ \hline  \hline
MSE & 1.0989 & 0.1333 & 0.4507 & 0.5211 & 0.1960 & 0.0630  \\
$95\%$ CI & 93.10 & 94.21 & 75.28 & 93.71 & 71.68 & 95.02  \\
Tr. Time & 3266.2 & 299.5 & 120.4 & 3022.0 & 23.1 & 108.6  \\ \hline
    \end{tabular}
    \caption{Comparison of the different models under study in the synthetic dataset.}
    \label{tab:results_toyproblem}
\end{table*}

\begin{figure*}
    \centering
    \includegraphics[width=\textwidth]{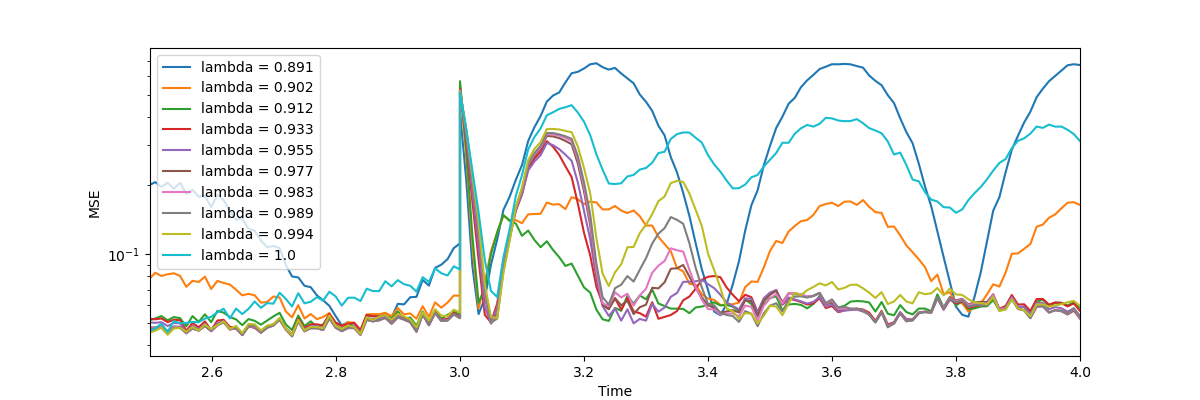} \\
    \caption{Evolution of the AGP MSE for different $\lambda$ values in the synthetic dataset.}
    \label{fig:MSE_toyproblem_lamb}
\end{figure*}

Table \ref{tab:results_toyproblem} clearly shows that the algorithm presenting the best performance is AGP, both in terms of MSE and estimation of the $95\%$ CI followed by OSGP and fast-AGP, the latter being a good choice if the computational burden is critical. Analyzing Figures \ref{fig:MSE_toyproblem} and \ref{fig:Pred_dist_toyproblem}, we can check as all the models work properly in the first part of the signal (when the frequency is low and only have to follow a change of amplitude), but when the signal frequency changes all the methods under study except AGP fail. For example, the fast-AGP is not able to track the signal because it does not update the kernel and noise parameters, so it is not able to follow the signal variations.
w-SVGP and AGP-VSI (mainly the former) fail because they are window based and do not have a forgetting factor, so when the signal changes fast they are only able to learn the average behavior of the signal; if the window length were smaller, the methods fall short of information to properly converge. 
 WISKI tries to track the signal variations, but it struggles to capture the amplitude, and their estimations are delayed (see Fig. \ref{fig:Pred_dist_toyproblem} bottom right); the OSGP is the only reference algorithm able to track the signal changes, but its confidence interval estimation grows along the time,  every time being more pessimistic.

To complete this analysis, Figure \ref{fig:MSE_toyproblem_lamb} shows the evolution of the MSE in AGP for different values of $\lambda$ focusing this analysis on the region around the sinusoidal frequency change (from $2.5$ to $4$). We set up parameter $\lambda\leq 1$ so the most remote sample of the window has an attenuation $\lambda^T=10^{-v}$, where $v$ takes the values $0, 0.25, 0.5, 0.75, 1, 2$ and $3$.   

As we can see, the introduction of adaptive learning is critical to fast-track the signal transition and to match the high frequency of the new sinusoidal. If $\lambda = 1$ (no forgetting is applied), the AGP model has a performance similar to the baselines. Setting $\lambda$ to any of the other values, the algorithm adapts the model to the new signal frequency.  The tracking error decreases as $\lambda$ is reduced  from $1$ to $0.933$ (see the MSE differences in the interval from $t=3.2$ to $t=3.4$).

Figure 4 shows the evolution of the MSE in AGP for different values of $\lambda$, focusing this analysis on the region around the sinusoidal frequency change (from $2.5$ to $4$). We set up parameter $\lambda\leq 1$ so the most remote sample of the window has an attenuation $\lambda^T=10^{-v}$, where $v$ takes the values $0, 0.25, 0.5, 0.75, 1, 2$, $3$, $4$, $4.5$ and $5$.  

As we can see, the introduction of adaptive learning is critical to fast-track the signal transition and to match the high frequency of the new sinusoidal. If $\lambda = 1$ (no forgetting is applied), the AGP model has a performance similar to the baselines. Setting $\lambda$ to any of the other values, the algorithm adapts the model to the new signal frequency.  The tracking error decreases as $\lambda$ is reduced from $1$ to $0.912$ (see the MSE differences in the interval from $t=3.2$ to $t=3.4$). However, if $\lambda$ is too low ($0.9$ or lower), the algorithm forgets too fast and does not have enough information to learn the signal evolution.

\subsection{Load forecast}

The dataset used in this article is the archived data ISO-New England\footnote{The dataset is publicly available at \url{https://www.iso-ne.com/isoexpress/web/reports/load-and-demand/-/tree/zone-info.}}, from 2011 to 2020, including real-time demand of each one of the eight load zones of New England.  ISO-New England Inc. is a Regional Transmission Organization (RTO), an Independent System Operator that operates the bulk power electric system and transmission lines serving the territory of the six New England states: Connecticut (CT), Maine (ME), Northeast Massachusetts and Boston (NEMA), New Hampshire (NH), Rhode Island (RI), Southeast Massachusetts (SEMA),  Vermont (VT), and Western/Central Massachusetts (WCMA).

The input patterns consist of 24 hourly power measures of a single day to predict every hour of the next day, and we have used an observation window of 90 days ($T=90$). Assuming that the most remote sample of the window has to be attenuated by a factor of 0.1, this leads to $\lambda =  0.9747$. The initial model has also been trained with $90$ data, and we have used $M=20$ inducing points, but the fast-AGP and AGP algorithms select between 8 and 10 inducing points in all scenarios.

As the model is using $24$ input variables, the WISKI algorithm cannot work in such a high space, so we include a preprocessing step (as suggested in \cite{stanton2021kernel}) to project the data into a two-dimensional space with a one-layer neural network consisting in a set of linear weights, a batch normalization and an output hyperbolic tangent to transform the data into the $[-1,1]$ scale. This way, the inducing point grid is predefined in this dimensional space with $400$ points equally lying in the grid $[-1,1] \times [-1,1] $. Using less dense grids  turns into poor results.  

The OSGP reference approach has to be removed from this experiment since it has shown convergence errors in most regions, providing poor performance. These converge problems are due to the expression of its inference bound, which has to invert a matrix defined by the subtraction of two terms of a similar order, causing its  Cholesky decomposition, despite being computed with a robust implementation, to fail in many realizations.

\begin{figure*}[t]
    \centering
    \includegraphics[scale=0.5]{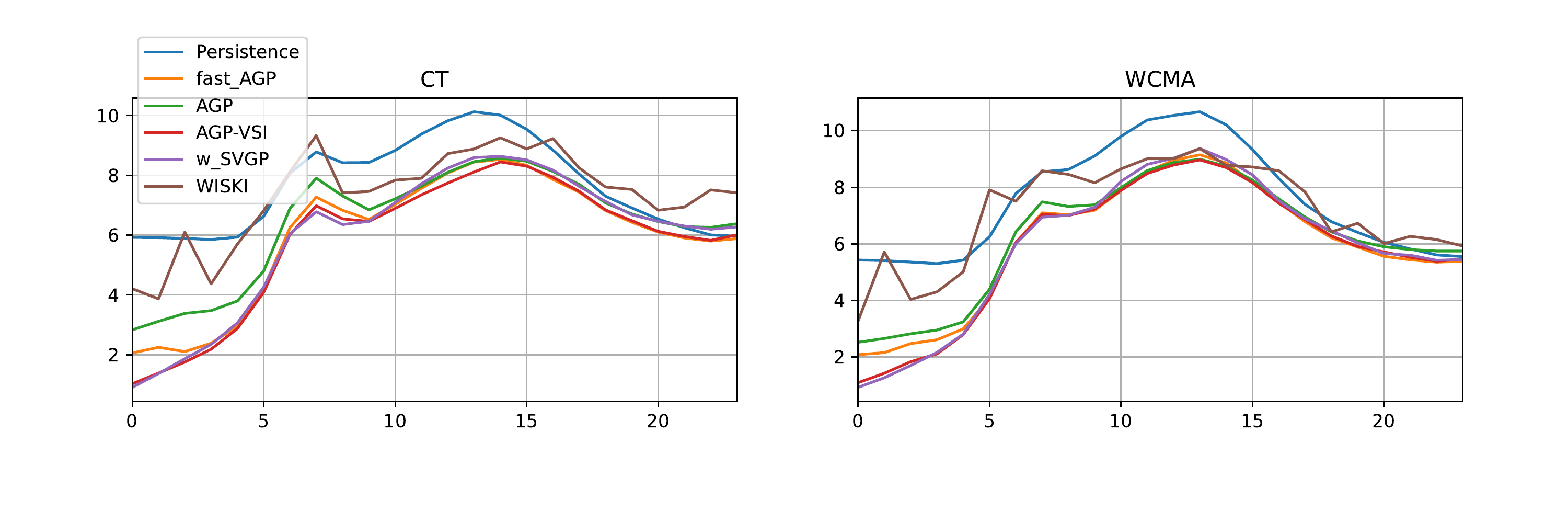}
    \caption{MAPE comparison of all algorithms for the power load experiment in the regions of Connecticut and West/Central Massachussets.}
    \label{fig:MAPE_all}
\end{figure*}

Figure \ref{fig:MAPE_all} shows the mean absolute percentage of error (MAPE) of the 24-hour prediction achieved by all methods in the regions of Connecticut (CT) and West/Central Massachusetts. In these results, we also include the performance of the \emph{persistence} approach, that is, an estimation given by the load consumption on the previous day at the same hour. The best MAPEs correspond to the fast-AGP and AGP algorithms, the WISKI algorithm shows similar performance to the simple persistence, and AGP-VSI and w-SVGP show performances close to AGP and fast-AGP ones. The results for all areas are summarized in Table \ref{tab:results_load}.

\begin{table*}[t]
    \centering
    \begin{tabular}{l|l||c|c|c|c|c|c|}
& & Persist. & w-SVGP &  WISKI & AGP-VSI & fast-AGP & AGP \\ \hline  \hline
 & MSE & 0.1282 & 0.0982 &  0.1202 & 0.0876 & 0.0876 & 0.0933  \\
CT & $95\%$ CI & -- & 89.91 &  90.22 & 89.82 & 90.70 & 96.16  \\
 & Tr. Time & -- & 5262.4 & 280.4 & 5459.9 & 36.6 & 196.0  \\ \hline
 & MSE & 0.0092 & 0.0062 &  0.0090 & 0.0112 & 0.0058 & 0.0066  \\
ME & $95\%$ CI  & -- & 91.96 &  91.44 & 92.78 & 93.88 & 96.70  \\
 & Tr. Time & -- & 5535.3 & 278.8 & 5519.2 & 60.5 & 198.5  \\ \hline
 & MSE & 0.0802 & 0.0593 &  0.0706 & 0.0535 & 0.0490 & 0.0533  \\
NEMA & $95\%$ CI  & -- & 89.74 &  89.14 & 90.01 & 89.73 & 96.20  \\
 & Tr. Time & -- & 5178.2 & 280.2 & 5162.3 & 38.7 & 197.7  \\ \hline
 & MSE & 0.0160 & 0.0107 &  0.0148 & 0.0121 & 0.0097 & 0.0107  \\
NH & $95\%$ CI  & -- & 91.50 &  90.46 & 91.21 & 92.38 & 96.63  \\
 & Tr. Time & -- & 5209.2 & 284.9 & 5107.9 & 54.6 & 196.8  \\ \hline
 & MSE & 0.0088 & 0.0099 &  0.0100 & 0.0134 & 0.0061 & 0.0069  \\
RI & $95\%$ CI  & -- & 90.75 &  89.42 & 92.26 & 88.00 & 96.35  \\
 & Tr. Time & -- & 5180.2 & 289.9 & 5154.2 & 65.1 & 196.3  \\ \hline
 & MSE & 0.0347 & 0.0280 &  0.0375 & 0.0281 & 0.0247 & 0.0263  \\
SEMA & $95\%$ CI  & -- & 90.19 &  91.33 & 89.65 & 89.15 & 95.84  \\
 & Tr. Time & -- & 5240.5 & 286.6 & 5105.7 & 45.6 & 196.2  \\ \hline
 & MSE & 0.0056 & 0.0042 &  0.0049 & 0.0044 & 0.0038 & 0.0040  \\
VT & $95\%$ CI  & -- & 92.58 &  92.58 & 93.77 & 97.14 & 97.34  \\
 & Tr. Time & -- & 5128.1 & 284.7 & 5186.4 & 54.2 & 196.1  \\ \hline
 & MSE & 0.0360 & 0.0255 &  0.0309 & 0.0240 & 0.0237 & 0.0248  \\
WCMA & $95\%$ CI  & -- & 89.84 &  92.00 & 89.90 & 93.15 & 96.84  \\
 & Tr. Time & -- & 5042.0 & 287.6 & 5290.0 & 37.9 & 194.4  \\ \hline
 & MSE & 0.0398 & 0.0303 &  0.0372 & 0.0293 & 0.0263 & 0.0282  \\
Average & $95\%$ CI  & -- & 90.81 &  90.82 & 91.17 & 91.77 & 96.51  \\
 & Tr. Time & -- & 5222.0 & 284.1 & 5248.2 & 49.2 & 196.5  \\ \hline
    \end{tabular}
    \caption{Comparison of the different models under study for the different load forecasting regions.}
    \label{tab:results_load}
\end{table*}

Analyzing the detailed results of Table  \ref{tab:results_load}, we can corroborate that fast-AGP and AGP show the best performance. Although fast-AGP gets the lowest MSE, AGP is the only one able to obtain an accurate $95\%$ CI estimation. This advantage of AGP in terms of the mean estimation can be due to a slight overfitting of fast-AGP. 

Regarding reference methods, w-SVGP, AGP-VSI and WISKI present an acceptable performance, but if we analyze the MAPE results (see Figure \ref{fig:MAPE_all}), we see that WISKI is quite unstable and in some time hours presents an error higher than the persistence approach. We think this punctual performance degradation is probably  because the predefined grid of inducing points is insufficient or is badly placed for some hour estimations.

However, when we analyze the confidence interval estimation, we observe that AGP is the only one that systematically provides an accurate estimate  since it is around 96\% (perhaps slightly conservative), but the remaining methods are around 90\%..

Finally, if we analyze the computational cost, we corroborate that w-SGP and AGP-VSI are not affordable, since their computational burden is $25$ times higher than AGP and 100 times higher than fast-AGP. WISKI presents a computational burden similar to the AGP one,  but fast-AGP is the fastest with a computation time $5.7$ times lower than WISKI. 

In an environment such as energy load forecast, where kernel and noise parameters do not need to be recalculated quickly, fast-AGP can be a good choice. However, if an accurate confidence interval estimation is needed, the AGP can be a better option. Even for these semi-stationary environments both models can be combined (for instance running AGP every ten days stamps to track the model parameters) and obtain better variance estimation and higher speed.

\subsection{Purchase prediction}
In this subsection, we analyze the performance of the model over two additional datasets related to sales prediction tasks. In particular, we have selected two open datasets:
\begin{itemize}
    
\item The Store Sales (SS) from a Kaggle Time Series Forecasting competition\footnote{https://www.kaggle.com/competitions/store-sales-time-series-forecasting}. In this case, the dataset provides the sale records of a grocery retailer in different categories. We have merged all the categories. The goal of the model is to predict the total amount of sales during the next day using the records of sales during the previous week.
    \item The Online Retail (OR) from the UCI repository\footnote{Avaliable at https://archive.ics.uci.edu/ml/datasets/Online+Retail}. Here, we aim to predict the revenues generated by total sales on each day using the incomings generated during the previous week.
\end{itemize}

In both datasets, we have used a similar configuration to the load forecasting problem, that is, we have considered an observation window of 90 days ($T=90$), which provides a $\lambda$ value of $  0.9747$ and for the fast-AGP and AGP algorithms we have used a maximum of $M=20$ inducing points. Again, for the WISKI algorithm, we have included a previous neural network with a hyperbolic tangent output to project the data into a two-dimensional space.

The final performance of the different methods is included in Table \ref{tab:results_purchase}. We have included the MSE of each model, their IC at 95\%, and the training times of each method. In these datasets, the OSGP algorithm only converged for the OR dataset, so their results are only provided for this case.

\begin{table*}[t]
    \centering
    \begin{tabular}{l|l||c|c|c|c|c|c|c|}
& & Persistance & w-SVGP & OSGP & WISKI & AGP-VSI & Fast-AGP & AGP \\ \hline  \hline
 & MSE & 0.4106 & 0.1744 & -- &  0.3877 & 0.1966 & 0.1868 & 0.2180  \\
 SS & IC 95 & -- & 87.63 & -- &  83.21 & 90.47 & 84.28 & 95.45  \\
 & Tr. Time & -- & 1911.6 & -- & 226.7 & 1833.5 & 39.6 & 297.7  \\ \hline
 & MSE & 3.9486 & 2.4789 & 2.5190 &  2.0593 & 2.2916 & 2.1920 & 2.2394  \\
 OR & IC 95 & -- & 84.39 & 89.76 &  89.27 & 87.32 & 93.17 & 94.15  \\
 & Tr. Time & -- & 240.6 & 30.5 & 29.3 & 232.0 & 3.0 & 20.7  \\ \hline
     \end{tabular}
    \caption{Comparison of the different models under study for the Store Sales (SS) and Online Retail (OR) datasets.}
    \label{tab:results_purchase}
\end{table*}

These results confirm the ones of the previous sections since, again, the proposed models tend to outperform the remaining algorithms providing the best trade-off of error, IC estimation, and reduced training time. Once again, the fast-AGP provides the best MSE with the faster implementation, but the AGP is the one that achieves the best estimation of the IC. In the OR dataset the WISKI approach stands out as the one providing the lowest MSE value, although it fails to achieve an accurate estimation of the IC 95.

\section{Conclusions}

Adaptive algorithms are used in signal processing and in these applications where the environment is non-stationary. We introduced an algorithm that implements an adaptive SGP for regression based on the VSGP introduced in \cite{titsias2009variational}, where a variational approach is used to maximize a variational lower bound of the marginal likelihood of the training regressors. 

To endow the VSGP with adaptive properties, the components of the log-likelihood are weighted with a factor that exponentially decays with time, so the information included in the log-likelihood is conveniently forgotten in a way similar to what is done in the RLS algorithm. To make this model efficient, the algorithm used for the inference is formulated in two alternative ways that make it adequate for online learning. In one of the approaches,  the noise and kernel parameters are frozen after initial training, and only the inducing points are adapted, to obtain the fast-AGP, which combines good performance and low computational burden. The second one, called AGP, updates all parameters in an online way, thus improving its capabilities by increasing the computational burden. In all cases, the computational burden is lower than the state-of-the-art approaches. To limit the number of inducing points, we propose a relevance criterion,  based on the change of the marginal likelihood, to automatically add a new sample as an inducing point or remove an existing inducing point. 

The performance of the algorithm has been tested in two problems. The first one is an artificial problem where a sinusoidal signal has to be tracked (with one-step ahead prediction). This signal shows a sudden frequency change that is properly tracked by the proposed algorithms but not by the state-of-the-art algorithm used in the comparison. The fast-AGP shows the second best performance, while the AGP shows the best performance. A test with different values of the forgetting factor $\lambda$ shows the importance of the forgetting factor, and it also shows the robustness of the algorithm regarding the parameter choice. 

A second experiment consists of electric load forecasting with the ISO New England load databases. The results show that both AGP and fast-AGP show the best performance in mean. Besides, the AGP is the only one that achieves an accurate 95\% CI, while the rest of the approaches show a 91\% of samples inside the estimated 95\% CI. This is important in these applications, where it is useful to know what is the confidence interval of the prediction to decide whether to trust the prediction.

\appendices

\section{Derivation of the $\lambda$-dependent log likelihood}\label{appendix:lambda_bound}

In this appendix, we derive the different expressions of Section \ref{sect:adapt_SGP} to demonstrate the solution for the adaptive version of the VSGP formulation.

Let's start including $\lambda$ into $\log G(\bbf_u, \by) $ 
\begin{equation}\label{eq:lambda_loglik_Appendix}
\begin{split}
& \log G^{\lambda}(\bbf_u, \by) = \sum_{t'=1}^t  \lambda^{t-t'} \log G(\bbf_u, y_{t'})\\
=  &\sum_{t'=1}^t  \lambda^{t-t'}  \log \mathcal{N} \left( y_{t'} |
\bk_{ut'}^{\top} \bK_{uu}^{-1} \bbf_u, \sigma^2 \right) \\
& -\sum_{t'=1}^t \frac{\lambda^{t-t'} }{2\sigma^2} \left( k_{t't'} -\bk_{ut'}^{\top} \bK_{uu}^{-1} \bk_{ut'} \right) 
\end{split}
\end{equation}
and, now, focusing on the term of the first sum and defining $\alpha_{t'} =\bk_{ut'}^{\top} \bK_{uu}^{-1} \bbf_u$, we can be reformulated this term as:
\begin{equation}
\begin{split}
&  \lambda^{t-t'}  \log \mathcal{N} \left( y_{t'} | \alpha_{t'}, \sigma^2 \right) \\
&  = -\frac{\lambda^{t-t'}}{2}  \left(  \log \left(2\pi\sigma^2\right) + \frac{1}{\sigma^2} \left(y_{t'} - \alpha_{t'}\right)^2  \right) \\
&  = -\frac{1}{2} \lambda^{t-t'} \log \left(2\pi\sigma^2\right) - \frac{1}{2}  \frac{\lambda^{t-t'}}{\sigma^2} \left(y_{t'} - \alpha_{t'}\right)^2 \\
&  ~~~ +  \frac{1}{2} \log\left(\frac{2\pi\sigma^2}{\lambda^{t-t'}}\right) - \frac{1}{2} \log\left(\frac{2\pi\sigma^2}{\lambda^{t-t'}}\right)  \\
&=\log \mathcal{N} \left( y_{t'} | \alpha_{t'},  \frac{\sigma^2}{\lambda^{t-t'}}  \right) \\
&  ~~~ -\frac{1}{2} \lambda^{t-t'} \log\left(2\pi\sigma^2\right) + \frac{1}{2} \log\left(\frac{2\pi\sigma^2}{\lambda^{t-t'}}\right).
\end{split}
\end{equation}
Now, we including this term into \eqref{eq:lambda_loglik_Appendix}, we get that 
\begin{equation}
\begin{split}
   &\log  G^{\lambda}(\bbf_u, \by) = \log \mathcal{N} \left( \by | \bK_{xu} \bK_{uu}^{-1} \bbf_u, \sigma^2 {\bLambda}^{-1} \right) \\ 
   &  - \frac{1}{2} \sum_{t'=1}^t  \left( 
    \log \lambda^{t-t'} + (\lambda^{t-t'}-1) \log(2\pi\sigma^2) \right) \\
    & - \frac{1}{2} \sum_{t'=1}^t   \frac{\lambda^{t-t'}}{\sigma^2}\left( k_{t't'} -\bk_{ut'}^{\top} \bK_{uu}^{-1} \bk_{ut'} 
    \right).
\end{split}
\end{equation}
where $\bLambda $ is a $t\times t$ diagonal matrix with $\bLambda_{t't'} =  \lambda^{t-t'}$.

Once $\log  G^{\lambda}(\bbf_u, \by)$ is defined, we can obtain the $\lambda$ dependent marginal likelihood or variational bound. For this purpose, let's remind that $p(\bbf_u)$ is 

\begin{equation}
p(\bbf_u) = \mathcal{N} \left(\bbf_u | \bZero, \bK_{uu}\right)
\end{equation}

and marginalizing $\log  G^{\lambda}(\bbf_u, \by)$ respect to $\bbf_u$
\begin{equation}
\begin{split}
    &F_V^{\lambda}(\bU)= \log \mathcal{N} \left( \by | \bZero, \sigma^{2}\bLambda^{-1} + \bK_{xu} \bK_{uu}^{-1} \bK_{ux}   \right) \\ 
    &  - \frac{1}{2} \sum_{t'=1}^t \left(  
    \log \lambda^{t-t'}  +   (\lambda^{t-t'}-1)  \log(2\pi\sigma^2) \right) \\
    & - \frac{1}{2}\sum_{t'=1}^t  \frac{\lambda^{t-t'} }{\sigma^2}\left( k_{t't'} -\bk_{ut'}^{\top} \bK_{uu}^{-1} \bk_{ut'} 
    \right) 
\end{split}
\end{equation}

and simplifying it, by removing the constant terms, we get
\begin{equation}
\begin{split}
    &F_V^{\lambda}(\bU)= \log \mathcal{N} \left( \by | \bZero, \sigma^{2}\bLambda^{-1} + \bK_{xu} \bK_{uu}^{-1} \bK_{ux}  \right) \\ 
    &  - \frac{1}{2}\sum_{t'=1}^t  (\lambda^{t-t'}-1)  \log(2\pi\sigma^2) \\
    & +\frac{1}{2}\sum_{t'=1}^t  \frac{\lambda^{t-t'} }{\sigma^2}\left( k_{t't'} -\bk_{ut'}^{\top} \bK_{uu}^{-1} \bk_{ut'} 
    \right)
\end{split}
\end{equation}

\section{An Adaptive SGP with Stochastic Variational Inference}\label{sec:VSI_VSGP} 

An alternative optimization of \eqref{eq:F_V_1} can be carried out by means of Stochastic Variational Inference (SVI) \cite{hoffman2013stochastic}. For this purpose, we first rewrite the true marginal bound \eqref{eq:F_V_1} as an Evidence Lower Bound (ELBO) 
\begin{equation}
F_{V}(\bU, q(\bbf_u)) = \mathbb{E}_{q(\bbf_u) } \left[ \log {G(\bbf_u, \by)p(\bbf_u)} -\log {q(\bbf_u)} \right] 
\end{equation}
Secondly, we take into account that  $G(\bbf_u, \by)$ factorizes with respect to the data, 
\begin{equation}
\begin{split}
  \log G(\bbf_u, \by) =&  \sum_{n=1}^N \log G(\bbf_u, y_n)\\
  =& \sum_{n=1}^N  \log \mathcal{N} \left( y_n | \bk_{nu} \bK_{uu}^{-1} \bbf_u, \sigma^2 \right) \\
   & - \sum_{n=1}^N \frac{1}{2\sigma^2} \left( k_{nn} - \bk_{un}^{\top} \bK_{uu}^{-1} \bk_{un}  \right)
  \end{split}
\end{equation}
where $\bk_{nu}$ is a vector with the kernel products of $\bx_n$ with the inducing points and $k_{nn}$ is the kernel of $\bx_n$ with itself. 

Finally, a stochastic expression of the marginal bound can be obtained as
\begin{equation}\label{eq:F_SVI}
F_{SVI}(\bU, q(\bbf_u)) = \mathbb{E}_{q(\bbf_u) } \left[ \sum_{n=1}^N \log {G(\bbf_u, y_n) p(\bbf_u)} -\log {q(\bbf_u)} \right].
\end{equation}

An interesting point of expression \eqref{eq:F_SVI} relies on the fact that the bound over the marginal likelihood is now expressed as a sum over the training data, which facilitates endowing this formulation with adaptive capabilities by weighting each data term with forgetting factor $0<\lambda<1$. This way, taking into account that we are in a time-dependent scenario at time $t$ and including $\lambda$ into \eqref{eq:F_SVI}  new bound becomes

\begin{equation}\label{eq:F_SVI_lambda}
\begin{split}
& F_{SVI}^{\lambda}(\bU, q(\bbf_u))   \\
& = \mathbb{E}_{q(\bbf_u) } \left[  \sum_{t'=1}^t \lambda^{t-t'} \log {G(\bbf_u, y_{t'}) p(\bbf_u)} -\log {q(\bbf_u)} \right].
\end{split}
\end{equation}

By maximizing this lower bound, the inducing points $\bU$ can be updated at every instant, as well as the variational parameters $\bmu$ and $\bA$, the kernel parameters, and the noise variance. Besides, this can be easily carried out with probabilistic programming packages, such as Pyro \cite{bingham2019pyro}, which samples from the variational distribution $q(\bbf_u) $ to approximate this bound with a computational cost of $\mathcal{O}(NM^2)$.
If we want to reduce this computational burden or the required memory, we can work with a data window of length $T$ covering training data from instant $t-T+1$ to $t$. Thus, the sum over the data in Eq. \eqref{eq:F_SVI_lambda} can be replaced by a sum from $t-T+1$ to $t$ to obtain the values of $\bU$ $\bmu$ and $\bA$ at instant $t$. During the next instant, $t+1$, the sliding window advances one position and the parameters are updated in a stochastic fashion using the previous solution as the starting point. This way, the computational burden is $\mathcal{O}(TM^2)$.

So, including the forgetting factor $\lambda$, this formulation provides an adaptive SGP formulation, as experimental results show (see Section \ref{sec:experiments}). However, the final computational burden of this model is going to be similar to retraining an SGP in each iteration since we are going to need many iterations of the stochastic optimization to make the algorithm converge to an adequate solution. Although this number of iterations can be reduced by using the optimum values of the parameters at instant $t-1$ to compute its new value at time $t$, it is not low enough since the stochastic nature of the model forces us to sample in several iterations over the variational distribution to obtain a workable solution.

\bibliography{references}

\section*{Biography Section}
 

% \bf{If you include a photo:}

\begin{IEEEbiography}[{\includegraphics[width=1in,height=1.25in,clip,keepaspectratio]{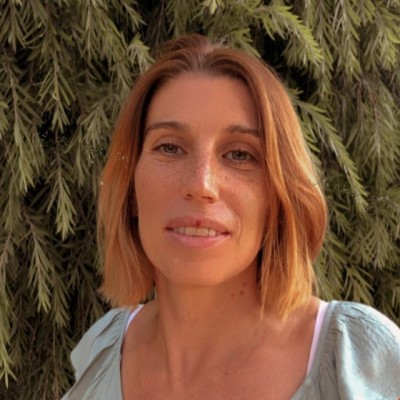}}]{Vanessa Gómez-Verdejo} received the Engineering degree in 2002 from Universidad Politécnica de Madrid. In 2007, she obtained a Ph.D. from Universidad Carlos III de Madrid, where she is currently Associated Professor. Her research interests are focused on machine learning, mainly, on probabilistic machine learning and feature selection methods and their use in health applications \url{http://vanessa.webs.tsc.uc3m.es/}.
\end{IEEEbiography}

\begin{IEEEbiography}[{\includegraphics[width=1in,height=1.25in,clip]{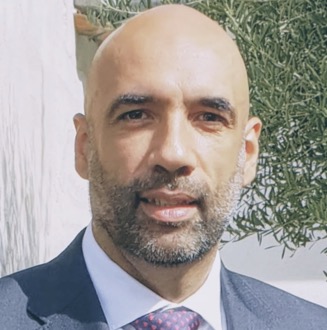}}]{Emilio Parrado-Hernández} Received an Engineering Degree from Universidad de Valladolid, Spain (1999) and a PhD in Communication Technologies from Universidad Carlos III de Madrid (2003). He currently is an Associate Professor at Universidad Carlos III de Madrid. From 2014 till 2019 Emilio worked as Senior Machine Learning Expert in  Advanced Analytics and Algorithmic Trading at BBVA. His research interests include machine learning, mostly kernel methods, and its application in finance and health.
\end{IEEEbiography}

\begin{IEEEbiography}[{\includegraphics[width=1in,height=1.25in,clip]{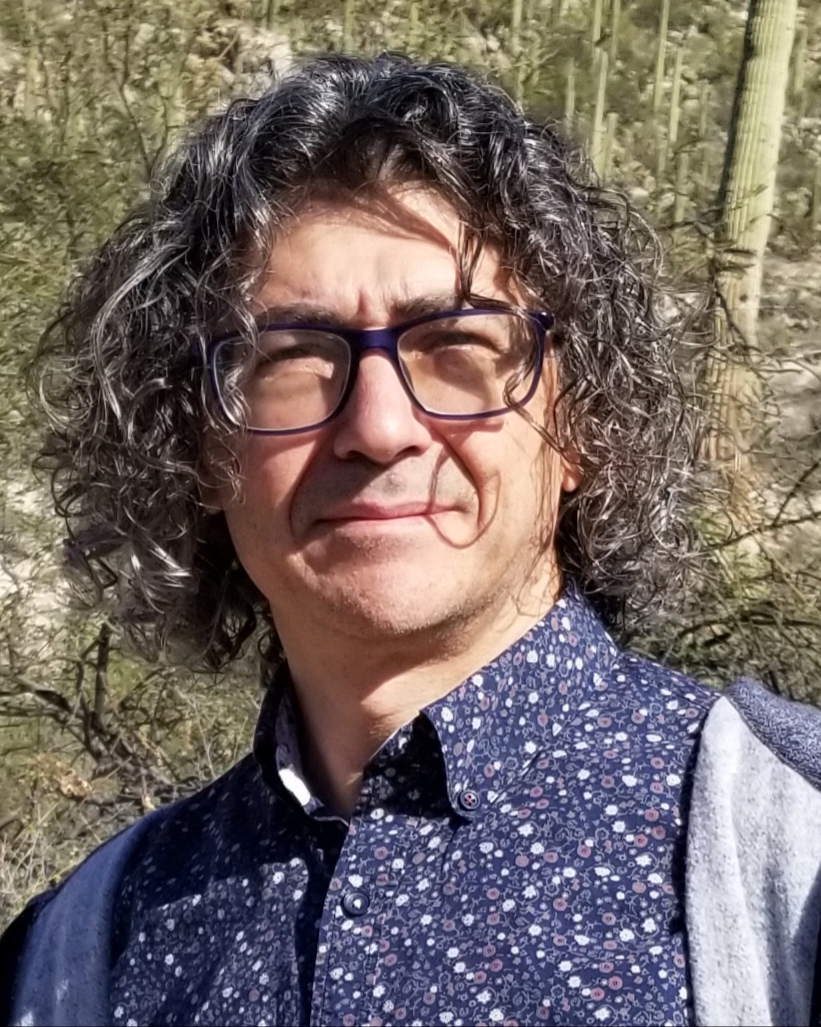}}]{Manel Martínez-Ramón} Received an Engineering Degree from Universitat Polit\`ecntica de Catalunya, Spain (1994) and a PhD in Communication Technologies from Universidad Carlos III de Madrid (1999). He is a professor at the Department of Electrical and Computer Engineering of the University of New Mexico, where he holds the King Felipe VI Endowed Chair in Information Sciences. His research activity is focused on Machine Learning and its applications to communications, smart grids, and complex systems, with emphasis on particle accelerators.
\end{IEEEbiography}

\end{document}